# Learning Bayesian Networks with Local Structure

**Nir Friedman**
Stanford University
Dept. of Computer Science
Gates Building 1A
Stanford, CA 94305-9010
nir@cs.stanford.edu

**Moises Goldszmidt***
SRI International
333 Ravenswood Way, EK329
Menlo Park, CA 94025
moises@erg.sri.com

## Abstract

In this paper we examine a novel addition to the known methods for learning Bayesian networks from data that improves the quality of the learned networks. Our approach explicitly represents and learns the *local structure* in the *conditional probability tables* (CPTs), that quantify these networks. This increases the space of possible models, enabling the representation of CPTs with a variable number of parameters that depends on the learned local structures. The resulting learning procedure is capable of inducing models that better emulate the real complexity of the interactions present in the data. We describe the theoretical foundations and practical aspects of learning local structures, as well as an empirical evaluation of the proposed method. This evaluation indicates that learning curves characterizing the procedure that exploits the local structure converge faster than these of the standard procedure. Our results also show that networks learned with local structure tend to be more complex (in terms of arcs), yet require less parameters.

## 1 Introduction

In recent years there has been a growing number of interesting results in the literature on learning Bayesian networks from data. Most of these results focus on learning the *global* structure of the network; that is, the edges of the directed acyclic graph that describes the independencies embodied by the network. Once this structure is fixed, learning the parameters in the *Conditional Probability Tables* (CPT) is usually solved by estimating a locally exponential number of parameters from the data. In this paper we propose the use of *local structures* for representing the CPTs and introduce the methods and algorithms for learning these structures as part of the process of learning the network. Using these structures we can model various degrees of complexity in the CPT representations. As we will show this considerably improves the quality of the learned networks.

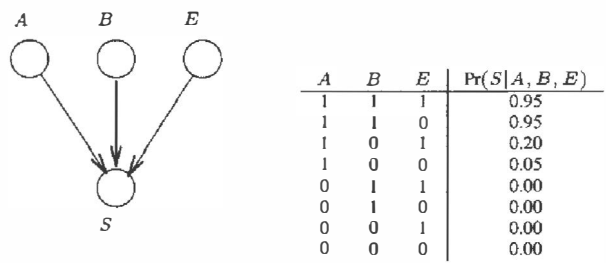

Figure 1: A simple network structure and the associated CPT for node $S$.

A Bayesian network represents a probability distribution whose parameters are specified by a set of CPTs. Each node in the network has an associated CPT that describes the conditional probability distribution of that node given the different values for its parents. In its most naive form, a CPT is encoded using a tabular representation which is locally exponential on the number of parents of a node: each assignment of values to the parents of a node requires the specification of a conditional distribution over that node. Thus, for example, consider the simple network in Figure 1, where the nodes $A$, $B$, $E$ and $S$ correspond to the events "alarm armed," "burglary," "earthquake" and "loud alarm sound," respectively. Assuming that all variables are binary, a naive tabular representation of the CPT for $S$ will require 8 parameters, one for each possible state of the parents. One possible quantification of this CPT is given in Figure 1. Note however, that when the alarm is not armed, (i.e., $A = 0$), the probability of $S = 1$ is zero, regardless of the values $B$ and $E$. Thus, the interaction between $S$ and its parents is simpler than the 8-way situation that is assumed in the naive representation of the CPT.

The locally exponential size of the naive representation of the CPTs is a major problem in learning Bayesian networks. As a general rule, learning many parameters is a liability, since a large number of parameters requires a large training set to be assessed reliably.[1] Thus, in general, learning procedures encode a bias against structures that involve many parameters. For example, given a training set with instances sampled from the network of Figure 1, the learning procedure might choose a simpler network structure over that of the original network. Using the naive tabular representation, the CPT for $S$ requires 8 parameters. However, a network with only two parents for $S$, say $A$ and $B$, would require only 4 parameters. Thus, for a small training set, such a network may be preferred, even though it ignores the effect of $E$ on $S$. The point of this example is to illustrate that by taking into account the number of parameters, the learning procedure may penalize a large CPT even if the interactions between the node and its parents are relatively benign. Our strategy is to alleviate this problem by explicitly representing the *local structure* of the CPT. This enables the learning procedure to consider each CPT according to the "real" number of parameters it actually requires to represent the conditional probability distribution, rather than the maximal possible number it might use with a naive representation. In other words, this explicit representation of local structure in the network's CPT allows us to adjust the penalty incurred by the network to reflect the real complexity of the interactions described by the network.

---
[1] This issue is related to the problem of induced models overfitting the training data.

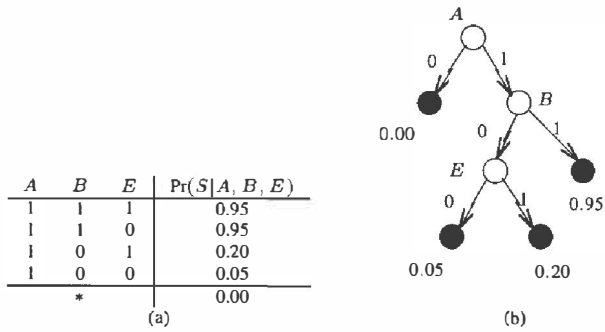

Figure 2: Example of the two representations of the local CPT structure. Part (a) shows a *default table*, and Part (b) shows a *decision tree*.

In this paper we examine two possible representations of the *local structure* of CPTs, and the methods for learning them. These representations, shown in Figure 2, will, in general, require fewer parameters than a naive representation. Part (a) in Figure 2 describes a *default table*, which is similar to the usual tabular representation, except that it does not list all of the possible values of $S$'s parents. Instead it provides a *default* probability assignment to all the values of the parents that are not explicitly listed. In this example, the default table uses 5 parameters as opposed to the original 8. Part (b) describes another possible representation based on *decision trees*. Each leaf in the decision tree describes a probability for $S$. The internal nodes of the tree encode possible values of $S$'s parents. In our example, the tree captures the additional structure that whenever $B = 1$ and $A = 1$, the probability of $S$ is the same regardless of the state of $E$. Thus, it requires 4 parameters instead of 8.

Our hypothesis is that incorporating local structure representations into the learning procedure leads to two important improvements in the quality of the induced models.

First, the parameters are more reliable. Since these representations usually require less parameters, the frequency estimation for each parameter takes a larger number of samples into account and thus they are more robust. Second, the global structure of the directed acyclic graph is a better approximation to the real (in)dependencies in the data. The use of local structure enables the learning procedure to explore networks that would have incurred an exponential penalty and thus would have not been taken into consideration. We cannot stress enough the importance of this last point. Finding better estimates of the parameters for a global structure that makes unrealistic independence assumptions will not overcome the deficiencies of the model. Thus, it is crucial to obtain a good approximation of the global structure. The experiments described in Section 5 validate this hypothesis. Moreover, the results in that section show that the use of local representations for the CPTs have a significant impact on the learning process itself. It translates into a faster learning to the distribution in the data. In other words, the learning procedures require fewer data samples in order to induce a network that better approximates the target distribution.

The main contributions of this paper are twofold. The first is the formulation of the hypothesis introduced above, which uncovers the benefits of having an explicit local representation for CPTs. The second is the empirical investigation that validates this hypothesis. In the process we also derive, in a principled manner, an MDL metric and algorithms for learning the local representations. In addition, we discuss the necessary modifications to the Bayesian metric of [Heckerman, Geiger, and Chickering 1995].

We are certainly not the first to suggest local structure for the CPTS. Such structures have been often used in knowledge acquisition from experts; the *noisy-or* gate and its generalizations are well known examples [Heckerman and Breese 1994; Pearl 1988; Srinivas 1993]. In the context of learning, it has been noted by several authors that CPTs can be represented using logistic regression, noisy-ors, neural networks and decision trees [Buntine 1991b; Diez 1993; Musick 1994; Neal 1992; Spiegelhalter and Lauritzen 1990]. With the exception of Buntine, these authors have focused on the case where the network structure is fixed in advance, and motivate the use of local structure for learning reliable parameters. Buntine does not limit his investigations to the case of a fixed structure, yet the advantages he foresees are the same ones that motivated the introduction of local structure in knowledge acquisition tasks. To the best of our knowledge, the benefits that relate to a more accurate induction of the global structure of a network have been unknown in the literature prior to this paper.

This paper is organized as follows: Section 2 reviews the definition of Bayesian networks, and the derivation of the *minimum description length* (MDL) score for learning Bayesian networks. Section 3 formally derives the MDL score for default tables and decision trees, and Section 4 describes the procedures for learning these structures. Section 5 presents the experimental results, and we conclude with a discussion and summary in Section 6. Appendix A describes the modifications needed for adapting the Bayesian scoring metric to networks with local structure.

## 2 Learning Bayesian Networks

Consider a finite set $\mathbf{U} = \{X_1, \ldots, X_n\}$ of discrete random variables where each variable $X_i$ may take on values from a finite domain. We use capital letters, such as $X, Y, Z$, for variable names and lowercase letters $x, y, z$ to denote specific values taken by those variables. The set of values $X$ can attain is denoted as $Val(X)$, the cardinality of this set is denoted as $||X|| = |Val(X)|$. Sets of variables are denoted by boldface capital letters $\mathbf{X}, \mathbf{Y}, \mathbf{Z}$, and assignments of values to the variables in these sets will be denoted by boldface lowercase letters $\mathbf{x}, \mathbf{y}, \mathbf{z}$ (we use $Val(\mathbf{X})$ and $||\mathbf{X}||$ in the obvious way). Let $P$ be a joint probability distribution over the variables in $\mathbf{U}$, and let $\mathbf{X}, \mathbf{Y}, \mathbf{Z}$ be subsets of $\mathbf{U}$. $\mathbf{X}$ and $\mathbf{Y}$ are *conditionally independent* given $\mathbf{Z}$ if for all $\mathbf{x} \in Val(\mathbf{X}), \mathbf{y} \in Val(\mathbf{Y}), \mathbf{z} \in Val(\mathbf{Z})$, $P(\mathbf{x} \mid \mathbf{z}, \mathbf{y}) = P(\mathbf{x} \mid \mathbf{z})$ whenever $P(\mathbf{y}, \mathbf{z}) > 0$.

A *Bayesian network* is an annotated directed acyclic graph that encodes a joint probability distribution of a domain composed of a set of random variables. Formally, a Bayesian network for $\mathbf{U}$ is the pair $B = \langle G, \Theta \rangle$. $G$ is a directed acyclic graph whose nodes correspond to the random variables $X_1, \ldots, X_n$, and whose edges represent direct dependencies between the variables. The graph structure $G$ encodes the following set of independence assumptions: each node $X_i$ is independent of its non-descendants given its parents in $G$ [Pearl 1988].[2] The second component of the pair, namely $\Theta$, represents the set of parameters that quantifies the network. It contains a parameter $\theta_{x_i|\Pi_{x_i}} = P(x_i|\Pi_{x_i})$ for each possible value $x_i$ of $X_i$, and $\Pi_{x_i}$ of $\Pi_{X_i}$ (the set of parents of $X_i$ in $G$). $B$ defines a unique joint probability distribution over $\mathbf{U}$ given by:

$$P_B(X_1, \ldots, X_n) = \prod_{i=1}^{n} P_B(X_i|\Pi_{X_i}) = \prod_{i=1}^{n} \theta_{X_i|\Pi_{X_i}} \quad (1)$$

The problem of learning a Bayesian network can be stated as follows. Given a *training set* $D = \{\mathbf{u}_1, \ldots, \mathbf{u}_N\}$ of instances of $\mathbf{U}$ (i.e., each $\mathbf{u}_i$ is a value assignment to all variables in $\mathbf{U}$), find a network $B$ that *best matches* $D$. To formalize the notion of goodness of fit of a network with respect to the data, we normally introduce a scoring metric, and to solve the optimization problem we usually rely on heuristic search techniques over the space of possible networks [Heckerman 1995]. Several different metrics have been proposed in the literature. In this paper we focus our attention on the *Minimal Description Length* (MDL) score [Lam and Bacchus 1994]. This score is simple, very intuitive, and has proven to be quite effective in practice. Another scoring metric that has received much attention recently is the Bayesian scoring metric [Cooper and Herskovits 1992; Buntine 1991b; Heckerman, Geiger, and Chickering 1995]. We defer the discussion of this metric and its modification to learn networks with local structure to Appendix A.

The MDL principle [Rissanen 1989] has a simple motivation in universal coding. Suppose that we are given a set $D$ of instances which we would like to store and keep in our records. Naturally, we would like to conserve space and save a compressed version of $D$. To this end we need to find a suitable model for $D$ such that an encoder can take this model and produce a compact image of $D$. Moreover, as we want to be able to recover $D$, we must also store a version of the model used by the encoder to compress $D$. The description length of the data based on a model, and using a particular encoder, is then the length of the compressed data plus the representation size of the model itself. The MDL principle dictates that the optimal model is the one (from a particular class of interest) that minimizes the total description length.

The MDL principle is applied to learning Bayesian networks by taking a network to be the model for the data used by an encoder to produced a compressed version of $D$. The idea is as follows: a network $B$ assigns a probability to each instance of $\mathbf{U}$. Using these probabilities we can construct an efficient code. In particular, we use the Huffman code [Cover and Thomas 1991], which assigns shorter codes to frequent instances. The benefit of using the MDL as a scoring metric is that the best network for $D$ optimally balances the complexity of the network with the degree of accuracy with which the network represents the frequencies in $D$.

We now describe in detail the representation length required for the storage of both the network and the coded data. The MDL score of a candidate network is defined as the total description length. To store a network $B = \langle G, \Theta \rangle$, we need to describe $\mathbf{U}, G$, and $\Theta$:

To describe $\mathbf{U}$, we store the the number of variables, $n$, and the cardinality of each variable $X_i$. Note that since $\mathbf{U}$ will be the same for each candidate network, we will ignore the description length of $\mathbf{U}$ in the comparisons between networks. Yet, we will assume that this information is present in the encoding for the rest of the terms in the description length.

To describe the DAG $G$ it is sufficient to store for each variable $X_i$ a description of $\Pi_{X_i}$ (namely, its parents in $G$). This description consists of the number of parents followed by a list of the parents. Since we can encode each of these using $\log n$ bits, the description length of the graph structure is:

$$DL_{graph}(G) = \sum_i (1 + |\Pi_{X_i}|) \log n.$$

To describe the parameters in $\Theta$, we must store the parameters in each conditional probability table. For the table associated with $X_i$, we need to store $||\Pi_{X_i}||(||X_i|| - 1)$ parameters. The representation length of these parameters depends on the number of bits we use for each numeric parameter. The usual choice in the literature is $1/2 \log N$ (see [Friedman and Yakhini 1996] for a thorough discussion of this point). Thus, the encoding length of $X_i$'s CPT is

$$DL_{table}(X_i, \Pi_{X_i}) = \frac{1}{2}||\Pi_{X_i}||(||X_i|| - 1) \log N.$$

We turn our attention to the description length of the data.

---

[2]Formally there is a notion of minimality associated with this definition, but we will ignore it in this paper. See [Pearl 1988] for details.

Using the probability measure defined by $B$, we construct a Huffman code for the instances in $D$. In this code, the exact length of each codeword depends on the probability assigned to that particular instance. There is no closed-form description of this length. However, it is known [Cover and Thomas 1991] that when we choose longer coding blocks we can approximate the optimal encoding length in which the encoding of each $\mathbf{u}$ is $-\log P_B(\mathbf{u})$ bits. Thus, the description length of the data is simply:

$$DL_{data}(\Theta_{X_i|\mathbf{\Pi}_{X_i}}, D) = -\sum_{i=1}^{N} \log P_B(\mathbf{u}_i).$$

We can rewrite this expression in a more convenient form. Let $\hat{P}_D$ be the empirical probability measure induced by the data set $D$. More precisely, we define

$$\hat{P}_D(A) = \frac{1}{N} \sum_{i=1}^{N} 1_A(\mathbf{u}_i) \text{ where } 1_A(\mathbf{u}) = \begin{cases} 1 & \text{if } \mathbf{u} \in A \\ 0 & \text{if } \mathbf{u} \notin A \end{cases}$$

for all events of interest, i.e., $A \subseteq \mathit{Val}(\mathbf{U})$. Using (1) we can rewrite the representation length of the data as:

$$DL_{data}(\Theta_{X_i|\mathbf{\Pi}_{X_i}}, D) = -N \sum_{x_i, \mathbf{\Pi}_{x_i}} \hat{P}_D(x_i, \mathbf{\Pi}_{x_i}) \log \theta_{x_i|\mathbf{\Pi}_{x_i}} \quad (2)$$

Thus, the encoding of the data can be decomposed based on terms that are "local" to each CPT: these terms depend only on $\Theta_{X_i|\mathbf{\Pi}_{X_i}}$. Standard arguments show that:

**Proposition 2.1:** *If $\Theta_{X|\mathbf{\Pi}_X}$ is represented as a table, then the parameters values that minimize $DL_{data}(\Theta_{X|\mathbf{\Pi}_X}, D)$ are $\theta_{x|\mathbf{\Pi}_x} = \hat{P}_D(x|\mathbf{\Pi}_x)$.*

Thus, given a fixed network structure $G$, learning the parameters that minimize the description length is straightforward: we simply compute the appropriate long-run fractions from the data.

Assuming that we assign parameters in $\Theta$ in the manner prescribed by this proposition, we can rewrite in $DL_{data}(\Theta_{X|\mathbf{\Pi}_X}, D)$ in a more convenient way in terms of conditional entropy: $N \cdot H(X_i|\mathbf{\Pi}_{X_i})$, where $H(X|Y) = -\sum_{x,y} \hat{P}_D(x,y) \log \hat{P}_D(x|y)$ is the *conditional entropy* of $X$ given $Y$. This gives a nice information theoretic interpretation for representation of the data: it measures how many bits are necessary to encode the value of $X_i$ once we know the value of $\mathbf{\Pi}_{X_i}$.

Finally, the MDL score of a candidate network structure $G$, assuming that we choose parameters $\Theta$ as prescribed above, is defined as the total description length

$$DL(G, D) = DL_{graph}(G) + \sum_i DL_{table}(X_i, \mathbf{\Pi}_{X_i}) + N \sum_i H(X_i|\mathbf{\Pi}_{X_i}).$$

According to the MDL principle, we should strive to find the network structure that minimizes this description length. In practice, this is usually done by searching over the space of possible networks.

We remark that the MDL score we just described coincides with the *Bayesian Information Criterion* (BIC) of [Schwarz 1978] which is related to Bayesian learning methods (see Appendix A). Roughly speaking, BIC would score a Bayesian network $B$ with $\log P_B(D) - \frac{1}{2} \log N \dim(B)$, where $\dim(B)$ is the dimension of $B$, i.e., the number of parameters it embodies. If we assume that the samples in $D$ are sampled independently from the same distribution, then $\log P_B(D) = \sum_i \log P_B(\mathbf{u}_i)$. Thus the BIC score (which one attempts to maximize) is the negative of the MDL score (which one attempts to minimize), when we ignore the description of $G$.

## 3 Adding Local Structure

In the derivation of the MDL score in the previous section we used a simplistic encoding for representing the parameters $\Theta$. We assumed the usual representation of CPTs requiring a locally exponential number of parameters. To be precise, for each node $X_i$ we assumed that we need to encode $||\mathbf{\Pi}_{X_i}||(||X_i||-1)$ parameters. In practice, however, the relation between $X_i$ and its parents $\mathbf{\Pi}_{X_i}$ can be more benign, and some regularities can be then exploited to represent the same information with fewer parameters. In the example of Figure 1, the information in the CPT can be encoded with four parameters using the decision tree in Figure 2 (b), as opposed to the eight parameters required by the naive tabular representation in Figure 1.

In this section we focus on defining compact representations that will exploit these regularities in the relations between a node and its parents to provide a smaller representation. This is crucial since, as discussed above, the MDL metric tradeoffs the complexity of the network for accuracy in the representation. Thus, it has a bias for learning networks with smaller families. By using the exponential encoding we may be unduly penalizing nodes in a network with a large number of parents. Compact encodings, on the other hand, will take advantage of the simpler interaction between the node and its parents, and will allow the exploration of networks with large families, accurately scoring their fitness with the data.

This section describes the encoding of these compact representations, and the changes in the MDL scoring metric. In the next two sections we discuss in detail how these representations can be learned and present experimental results that show their effectiveness.

### 3.1 Default Tables

A *default table* is very similar to a standard tabular representation for a CPT, except that only a subset of the possible values of the parents of a node are explicitly represented as rows in the table. The values of the parents that are not explicitly represented as individual rows are mapped to a special row called the *default row*. The idea is that the probability of a node $X$ is the same for all the values of the parents that are mapped to the default row, therefore there is no need to represent these values separately using different entries. Consequently the number of parameters explicitly represented in a default table is less than the number of parameters in a naive representation for a CPT. In the example of Figure 2 all the values of the parents of $S$ where $A = 0$ (the alarm is not armed), are mapped to the default row in the table since the probability of $S = 1$ is equal to zero regardless of these values.

Formally, a default table is an object $\mathcal{D}$. We define $Rows(\mathcal{D})$ to be the set of rows in $\mathcal{D}$. The description length of a default table is quite simple. We start by encoding the number $k = |Rows(\mathcal{D})| - 1$ of rows in the table that explicitly represent specific values for $X$'s parents values. Then, we encode which of the $\binom{||\boldsymbol{\Pi}_{X_i}||}{k}$ sets of rows actually appear in the table. Finally we need to encode the parameters in the $k$ rows and in the default row. Thus the description length of a CPT using default table $\mathcal{D}$:

$$DL_{def}(X, \boldsymbol{\Pi}_X, \mathcal{D}) = \log ||\boldsymbol{\Pi}_X|| + \log \binom{||\boldsymbol{\Pi}_X||}{k}$$
$$+ \frac{1}{2}|k+1|(||X_i|| - 1)\log N.$$

Note that in the extreme case when all the rows in the conditional probability table have distinct values, the second term in this equation is equal to zero and the last term is equal to the original encoding presented in the previous section. The first term in this equation $\log ||\boldsymbol{\Pi}_{X_i}||$, encoding the actual number of rows in the table, represents a bookkeeping penalty that we pay for the additional flexibility of using default tables. Note however, that this term does not depend on the size $N$ of the training data and will be of little influence as $N$ grows.

We now turn to examine how the assumption that the CPT is represented by a default table $\mathcal{D}$ affects how well the model fits the data. We start by defining $\Gamma_\mathcal{D}$ to be the *characteristic* random variable of the table $\mathcal{D}$, which takes values in $Rows(\mathcal{D})$. The event $\Gamma_\mathcal{D} = r$ corresponds to the value(s) of parents associated with the row $r$. We now can refer to parameters in the table $\mathcal{D}$ as $\Theta_{X|\Gamma_\mathcal{D}}$.

**Proposition 3.1:** *If $\Theta_{X|\boldsymbol{\Pi}_X}$ is represented as a default table $\mathcal{D}$, then we can rewrite $DL_{data}(\Theta_{X|\boldsymbol{\Pi}_X}, D)$ as:*

$$-N \sum_{r \in Rows(\mathcal{D})} \sum_{x_i} \hat{P}_D(x_i, \Gamma_\mathcal{D} = r) \log \theta_{x_i|r}.$$

*Moreover, the parameter values for $\mathcal{D}$ that minimize $DL_{data}(\Theta_{X|\boldsymbol{\Pi}_X}, D)$ are*

$$\theta_{x_i|r} = \hat{P}_D(x_i|\Gamma_\mathcal{D} = r).$$

As in the case of a regular CPT representation, $DL_{data}$ is minimized when the parameters correspond to the appropriate frequencies in the training data. As consequence of this result, we get that for a fixed default table structure $\mathcal{D}$, the minimal representation length of the data is simply $N \cdot H(X|\Gamma_\mathcal{D})$. Thus, once again we get the information theoretic interpretation of $DL_{data}(\Theta_{X_i|\boldsymbol{\Pi}_{X_i}}, D)$ that measures how many bits are needed to represent $X$. This interpretation shows that the encoding of $X$ depends only on the values of $\Gamma_\mathcal{D}$. From the well known data processing inequality [Cover and Thomas 1991] we get that $H(X|\Gamma_\mathcal{D}) \geq H(X|\boldsymbol{\Pi}_X)$. This implies that a default table cannot fit the data better than a regular CPT. Nevertheless, the reduction in the number of parameters may compensate for the potential loss in information.

To summarize, the MDL score for a graph structure augmented with default table $\mathcal{D}_i$ for each $X_i$ is:
$DL_{def}(G, \mathcal{D}_1, \ldots, \mathcal{D}_n, D) =$
$DL_{graph}(G) + \sum_i \left( DL_{def}(X_i, \boldsymbol{\Pi}_{X_i}, \mathcal{D}_i) + N \cdot H(X|\Gamma_{\mathcal{D}_i}) \right).$

### 3.2 Decision Trees

In this context, a *decision tree* is a tree in which each internal node is annotated with a parent variable, outgoing edges from a particular node are annotated with the different values that the variable represented by that node can take, and leaves are annotated with a probability distribution over $X$. The process for retrieving the probability of $X$ given a value of its parents is as follows. We start at the root node and traverse the tree until we reach a leaf. At each internal node, we choose which subtree to traverse by testing the value of the parent annotating that node, and following the outgoing edge that corresponds to that value. Thus, suppose we would like to know $\Pr(S = 1|A = 1, B = 0, E = 1)$ in the tree shown in Figure 2(b). We follow the edge to the right subtree at $A$, since this edge is annotated with the value 1 for $A$. Similarly we follow the left edge on $B$ (annotated with 0) and again the right edge on $E$ till we reach the appropriate leaf.

Note that decision trees are more flexible than default tables in the sense that we can represent simpler interactions in a more compact manner. In general, a default table groups one set of values the parents can take (the ones that are not explicitly listed in the table) into a *partition*. A tree, on the other hand, can group several sets of such values, each one corresponding to a leaf in the tree. In our example, the leaf that corresponds to $A = 0$ groups 4 values of the parents of $S$, while the leaf that corresponds to $A = 1, B = 1$ groups two values of the parents (the other two leaves correspond each to a particular value of the parents).

For the formal definition of the description length, we will denote a tree as an object $\mathcal{T}$ which can either be a leaf or a composite tree. We introduce a function $Label(\mathcal{T})$ that returns the variable that is the root of $\mathcal{T}$, and a function $Sub(\mathcal{T}, v)$ that returns the sub-tree associated with the value $v$ of $Label(\mathcal{T})$. Given a tree $\mathcal{T}$ we define $Leaves(\mathcal{T})$ to be the set of leaves in $\mathcal{T}$.

The description length of a decision tree is composed of two parts: the description of the tree structure, and the description of the parameters in the leaves. For the description of the tree we follow the encoding proposed by Quinlan and Rivest [1989]. A tree is encoded recursively as follows: a leaf is encoded by a single bit with value equal to 0. The encoding of a composite tree starts with a bit set to the value 1, to differentiate it from a leaf, followed by a description of the associated test variable and the description of all the immediate sub-trees. The encoding of the test variable depends on the position of the node in the tree. At the root, the test variable can be any of $X$'s parents. However, since along a single path we test each variable at most once, we have smaller set of possibilities in deeper nodes. In general a node that is $k$ levels deep in the tree, would have $|\boldsymbol{\Pi}_X| - k$ possible candidates for the test variable. Thus, we need to store only $\log(|\boldsymbol{\Pi}_X| - k)$ bits in encoding. The

total description length of the tree structure is the following recurring formula:

- If $T$ is a leaf, then $DL_T(T, k, \Pi_X) = 1$.
- If $T$ is a composite tree with sub-trees $T_1, \ldots, T_m$, then

$$DL_T(T, k, \Pi_X) = 1 + \log(|\Pi_X| - k) + \sum_i DL_T(T_i, k+1, \Pi_X).$$

The description length of the parameters at the leaves is simply the number of leaves multiplied by $\frac{1}{2}(||(X_i) - 1||) \log N$. As noted in [Quinlan and Rivest 1989; Wallace and Patrick 1993] this encoding of the tree structure is suboptimal, especially when the tree has a high branching factor. In this paper, however, we use this simpler encoding since the description length is, in any case, dominated by the length of the parameters stored in the leaves (which depend on the sample size $N$).[3] Similar to the case of a default table there is a bookkeeping penalty for encoding the structure of the tree when compared with the naive encoding of a CPT. Once more, this penalty can be disregarded since it will be of little influence as $N$ grows, and it will only be relevant as the tree grows near a full tree.

Finally, the total encoding length of the CPT is:

$$DL_{tree}(X, \Pi_X, T) = DL_T(T, 0, \Pi_X) + \frac{1}{2}|Leaves(T)|(||X_i|| - 1) \log N.$$

For the description length of the data, we define the *characteristic* random variable $\Gamma_T$ that takes as values $Leaves(T)$. The event $\Gamma_T = l$ correspond to the state of the parents as represented by the labels on the edges that appear on the path from the root of $T$ to the leaf $l$. We get an analogous result to the one we had for default tables.

**Proposition 3.2:** *If $\Theta_{X|\Pi_X}$ is represented by a decision tree $T$, then we can rewrite $DL_{data}(\bullet_{X|\Pi_X}, D)$ as:*

$$-N \sum_{l \in Leaves(T)} \sum_{x_i} \hat{P}_D(x_i, \Gamma_T = l) \log \theta_{x_i|l}$$

*Moreover, the parameter values that minimize $DL_{data}(\Theta_{X|\Pi_X}, D)$ are*

$$\theta_{x_i|l} = \hat{P}_D(x_i|\Gamma_T = l).$$

We again get the expected information theoretic term for the encoding of the data using the best parameter values for a fixed tree structure $T$: $N \cdot H(X|\Gamma_T)$.

To summarize, the MDL score for a graph structure $G$ augmented with trees $T_i$ for each $X_i$ is:

$DL_{tree}(G, T_1, \ldots, T_n, D) =$
$DL_{graph}(G) + \sum_i (DL_{tree}(X_i, \Pi_{X_i}, T_i) + N \cdot H(X_i|\Gamma_{T_i}))$

---
[3]The more complex representation length of [Wallace and Patrick 1993] can be easily incorporated into our MDL score.

Table 1: Description of the three networks used to generate the sample data.

| Name | Description | $n$ | $||\mathbf{U}||$ | $|\Theta|$ |
|---|---|---|---|---|
| Alarm | A network by medical experts for monitoring patients in intensive care. | 37 | $2^{53.95}$ | 509 |
| CTS | A network developed by medical experts for diagnosing carpal tunnel syndrome. | 66 | $2^{74.85}$ | 525 |
| TJ | A network developed by domain experts for testing performance in Jet turbines | 34 | $2^{42.01}$ | 385 |

## 4 Learning Local Structures

In this section we describe an approach for learning the local structures (default tables or decision trees) given a particular global structure for the network. These procedures are applied independently to each CPT. Thus, in describing them we assume that we are given a variable $X$ and a set of parents $\Pi_X$, and the objective is to induce the local structure for this CPT. During the global learning process, these procedures will be called to find local structures for each new network candidate.

An important aspect of the scoring metrics we derived in the previous section is that they are decomposable. Thus, for example, the representation length of a tree is a sum of the representation lengths of the subtrees. Similarly, the scoring of the data using entropy is also decomposable. This was shown for the case of CPTs in Section 2, and it follows easily for the cases of default tables and decision trees. The decomposability property is crucial for developing incremental algorithms for learning the local structures.

For the learning of default tables we propose a simple minded greedy algorithm. We start with a trivial table with only the default row. Then, we iteratively add the row that minimizes the description length of the table and the data. This greedy expansion is repeated until no improvement in the score can be gained by adding another row.

We now turn to learning of decision trees. As pointed out by Quinlan and Rivest [1989], finding an optimal decision tree is apparently an intractable problem. There is a large body of work on procedures for the construction of decision trees (see for example [Quinlan 1993]). Here we adopt a straightforward approach outlined by Quinlan and Rivest.

The approach consists of two phases. In the first phase we "grow" the tree in a top-down fashion. We start with the trivial tree consisting of one leaf, and add branches to it until a maximal tree is learned. In the second phase we "trim" the tree in a bottom-up fashion.

To grow the tree we repeatedly replace a leaf with a subtree that has as root some parent of $X$, $Y$, and whose children are leaves; one for each value of $Y$. In order to decide on which parent $Y$ we should perform this *split* we compute the MDL score, i.e., $DL_{tree}$ defined above, of the tree associated with each parent, and select the parent which induces the best scoring tree. (The score can be computed in a *local* fashion by evaluating $H(X|Y)$ on the instances in the training data that are compatible with the path from the root of the tree to the node that is being split.) This procedure stops when either the node has no training instances associated with it, the value of $X$ is constant in the associated training set, or all the parents of $X$ have been tested along the path leading to that node.

The second phase is done by scanning the tree in a bottom-up manner. At each node we consider whether the representation length of the sub-tree rooted at that node is bigger or equal to the representation length of a leaf. If this is the case, then the sub-tree is trimmed and replaced with a leaf.

## 5 Experimental Results

The purpose of the experiments described in this section is to assess how the bias embodied by the different representations of conditional probability tables affects the learning behavior, and the quality of the induced models. To this end we collected data in the form of *learning curves* measuring the quality of the learned network as a function of the number of training samples, as well as different statistics regarding the number of parameters in the learned models.

We compared three different learning procedures which differ in the use of the local representation of conditional probabilities.

$G_{tab}$ uses the standard MDL score, as described in Section 2.

$G_{def}$ uses the MDL score based on default tables, as described in Section 3.1.

$G_{tree}$ uses the MDL score based on decision trees, as described in Section 3.2.

All three learning procedures use the same simple greedy search method for finding a candidate network. The starting point of the search is the empty network. We consider three possible types of operations on the candidate network: edge addition, edge removal and edge reversal. At each step, the procedure chooses the best operation among these, and applies it to the current candidate. (In $G_{def}$ and $G_{tree}$ this includes a search for the best local structure for the CPTs modified by each possible operation.) This process is repeated until the best modification does not improve the candidate's score. As expected, this hill-climbing search method is most likely to find a local minima instead of a global one. However, it has a reasonable behavior in practice (see [Heckerman, Geiger, and Chickering 1995]).

We tested the three learning procedures on data generated by three Bayesian networks described in Table 1.[4] From each of these networks we sampled training sets of 8 sizes—500, 1000, 2000, 4000, 6000, 8000, 12000, and 16000 instances—and run the learning procedures on them. In order to increase the accuracy of the results, we repeated the experiment with ten sets of training data.

### 5.1 Results

We are interested in comparing the use of structured representations in the learning procedures on three characteristics of the induced networks: number of instances needed versus overall quality, number of parameters learned (which indicates a measure of the robustness of these parameters), and the complexity of the network. We describe our results in turn.

---

[4]The Alarm network is well known and described in [Beinlich, Suermondt, Chavez, and Cooper 1989]. CTS and TJ were provided by Mark Peot of Knowledge Industries, after the variable names where appropriately sanitized.

To evaluate the overall quality of the network we compute the *cross-entropy* from the target distribution, that is, the one represented by the generating network, to the distribution represented by the learned network. This measure is defined as follows. Let $P$ be the target distribution and $Q$ the learned distribution. The cross entropy from $P$ to $Q$ is:

$$\sum_x P(x) \log \frac{P(x)}{Q(x)}.$$

This measure is the standard measure of distance in the Bayesian network learning literature [Cooper and Herskovits 1992; Heckerman, Geiger, and Chickering 1995; Lam and Bacchus 1994]. See [Friedman and Yakhini 1996] for a detailed discussion of this measure.

Figure 3 plots the learning curves for the three procedures described above. The figure displays the cross-entropy between the induced models and the generating model (vertical axis) versus the number of samples in the training data (horizontal axis). It was noted by [Friedman and Yakhini 1996], that, as a general rule, learning curves for these learning problems behave as a linear function of $\frac{\log N}{N}$. Thus, to facilitate comparisons we plot the learning errors scaled by $\frac{N}{\log N}$. Indeed, we observe that the resulting graphs are roughly constant. The dotted diagonal lines represent boundaries of constant error. All methods appear to converge to the target distribution (eventually they would intersect the dotted line of $\epsilon$ cross-entropy for all $\epsilon > 0$). However, both $G_{def}$ and $G_{tree}$ converge faster than $G_{tab}$. As a general rule we see a gap of $O(\frac{\log N}{N})$ between the error measure of $G_{tab}$, and $G_{def}$ and $G_{tree}$. The lines of constant error clearly indicate that as as $N$ grows larger, the number of samples $G_{tab}$ needs to reach an approximation compatible with $G_{def}$ (or $G_{tree}$) grows larger.

One surprising aspect of these results is the performance of default tables as compared to decision trees. In particular they are clearly better in small to medium (up to 8000) sample sizes. We suspect that this is due both to the low bookkeeping penalty in their encoding, and the fact that the greedy learning algorithm for default tables performs well. We note however, that decision trees perform better in larger sample sizes. For example, $G_{tree}$ performance improves as the sample size grows in CTS and TJ. This is due to the fact that many of the CPTs in CTS and TJ are represented using noisy-or and noisy-max [Heckerman and Breese 1994], which can be better approximated by trees (rather than by default tables). Another possible factor might be the way our trees handle multi-valued attributes. Whenever such an attribute is tested in a decision tree we must create many subtrees that fragment the sample into small groups. Default tables, on the other hand, can effectively group several values of multi-attribute variables into the "default" row. In future work we plan to address issue.

The next two experiments help in illustrating why the faster convergence of the methods using structured local representation. The first experiment is concerned with the number of parameters in the learned model, while the sec-

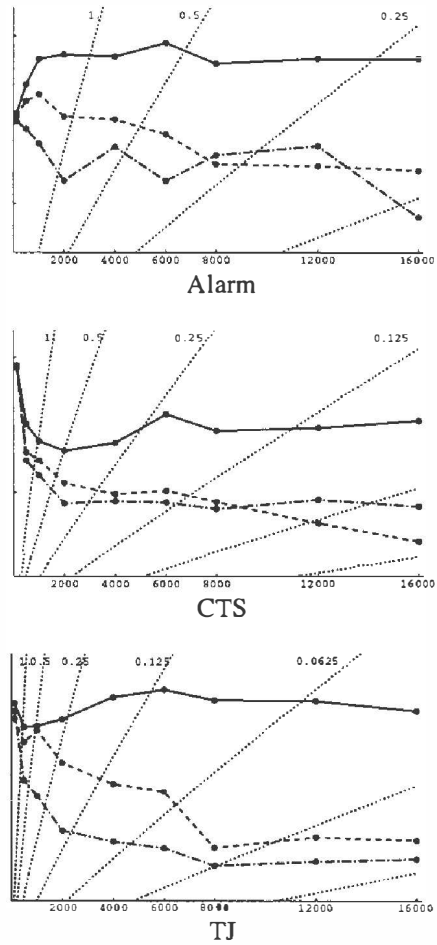

Alarm

CTS

TJ

Figure 3: Learning curves comparing the cross-entropy of networks learned using standard MDL (solid line) to networks learned with with trees (dashed line) and default tables (dot-dash line). The horizontal axis measures the number of samples, $N$. The vertical axis measures the error multiplied by $\frac{N}{\log N}$. The dotted diagonal lines are lines of constant error.

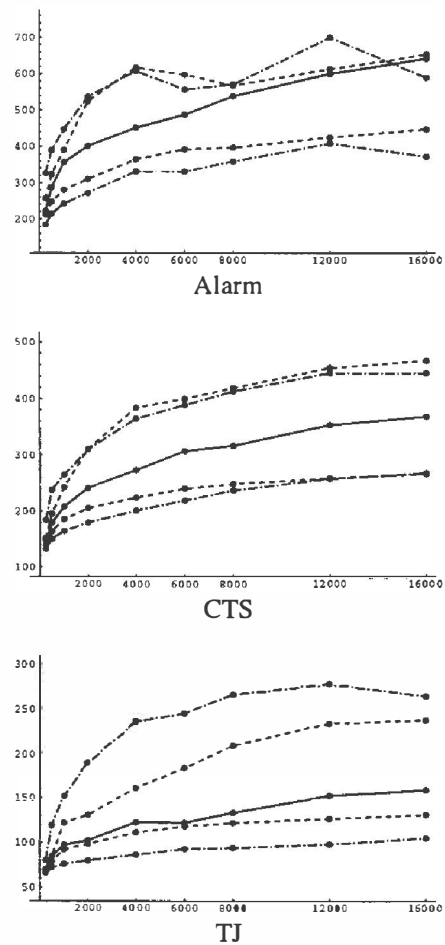

Alarm

CTS

TJ

Figure 4: Curves comparing the number of parameters and the complexity of the networks learned by standard MDL (solid line), using trees (dashed line) and using default tables (dot-dash line). The two measures are identical for standard MDL. Curves below the solid line measure the the actual number of parameters learned, and curves above the solid lines measure the number of parameters required had the learned network used tables to represent conditional probabilities.

ond is concerned with the *complexity* of these models. Results are depicted in Figure 4.

Generally, we say that a parameter is robust if it has low variance. The number of parameters can be used as a direct estimate of robustness of the learned parameters: For each random variable $X$, the parameters in $\Theta_{X|\Pi_X}$ are of the form $\hat{P}_D(X|E_1), \ldots, \hat{P}_D(X|E_k)$, where $E_1, \ldots, E_k$ are mutually disjoint and exhaustive events. The exact nature of $E_i$ depends on the representation of $\Theta_{X|\Pi_X}$, they can correspond to values of $\Pi_X$ (in the usual table representation), to leaves in the decision tree, or to rows in a default table. Since $E_1, \ldots, E_k$ are mutually disjoint and exhaustive, $\sum_j \hat{P}_D(E_j) = 1$. Thus, as $k$ grows larger, parameters are evaluated in smaller populations of instances, and thus run the risk of being less robust. Hence, as a general rule, for a fixed training data size, the fewer parameters in a model, the more reliable is their estimated values.

With respect to the complexity of the learned networks, we have that as a general rule, more complex networks make less assumptions of independence. A misguided assumption of independence introduces errors in the learned distribution that persist even if we can somehow obtain the optimal parameter values. Thus, we want to learn networks that are not much simpler than the target distribution requires.

Complexity of networks can be measured in various ways. We have chosen to measure it using the number of parameters we would have been required had we used the usual table representation of $\Theta_{X|\Pi_X}$. Since the number of parameters is exponential in the size of the family, this measure estimates how many "big families" are in the network. This measure is more exact than say counting edges, since it also takes into account the cardinality of parent variables. For example, if we take a variable $X$ in the network and add an edge directed to it from a variable with $k$ values, then the number of parameters needed for representing $\Theta_{X_i \Pi_X}$ is multiplied by $k$. Thus, an edge from a variable with larger cardinality incurs in a higher cost in terms of complexity.

Table 2: Approximation error for mixed structure/parameter learning methods. Rows describe the method used in learning the network structure, and columns describes the method used in learning the parameters. Results in part (a) of the table were learned from 10 sample sets of size 1000 from the CTS network and those in part (b) were learned from sample sets of size 4000 from the TJ network.

|  | Parameter Estimation Method ||||||
| --- | --- | --- | --- | --- | --- | --- |
|  | $G_{\text{tab}}$ | $G_{\text{tree}}$ | $G_{\text{def}}$ | $G_{\text{tab}}$ | $G_{\text{tree}}$ | $G_{\text{def}}$ |
| $G_{\text{tab}}$ | 0.954 | 0.903 | 0.890 | 0.161 | 0.134 | 0.127 |
| $G_{\text{tree}}$ | 0.992 | 0.855 | 0.848 | 0.176 | 0.108 | 0.092 |
| $G_{\text{def}}$ | 0.973 | 0.820 | 0.778 | 0.204 | 0.110 | 0.073 |
|  | (a) ||| (b) |||

Figure 4 describes the number of parameters and the complexity of the networks learned by the various methods. Note that $G_{\text{def}}$, the procedure that learns default tables, usually learns the smallest number of parameters, and at the same time induces models that are most complex. This combination reduces the variance of the estimated parameters, produces a more accurate representation of the (in)dependencies in the real distribution, and undoubtedly improves the cross-entropy of these networks. As expected, $G_{\text{tree}}$ learns less parameters than $G_{\text{tab}}$ and produces more complex networks. However, it usually learns more parameters then $G_{\text{def}}$.

In in attempt to isolate the improvement gained from learning a more complex structure and the improvement from learning fewer parameters we performed the following experiment. We took the structures learned by one method and learned the best parameters, holding the structure fixed, using the other methods. The results of this experiment can be found in Tables 2(a) and 2(b).[5] Note that once the global structure of the network is fixed, we can still obtain better approximations by learning local structures for the CPTs. This is evident by observing that the cross-entropy in these tables decreases as we traverse any row. In addition, both $G_{\text{tree}}$ and $G_{\text{tab}}$ learn better structures which lead to additional improvements in the approximations. It is interesting to observe that when we learn full CPTs for these larger (more complex) structures the error increases, since the parameters become unreliable. Thus, $G_{\text{tab}}$'s choice of small networks is, in this sense, justified.

In summary, these results validate our stated hypothesis. They suggest that the methods we propose find better (in terms of cross-entropy) models for two main reasons. First, they learn more complex structures. These structures do not make independence assumptions that do not hold in the underlying domain and thus, they reduce the error. Secondly, the learned networks contain fewer parameters. The estimation of these parameters is then more robust since it is based on a frequency over larger samples. They are closer to the actual probabilities in the underlying distribution. In this respect, both default tables and trees are flexible enough to regulate the number of learned parameters. If there are many instances, then more complex local structures can be learned (with more parameters). On the other hand, if there are few instances, then the local structure will be simpler and fewer parameters will be assessed.

## 6 Discussion

The main contribution of this paper is the introduction of structured representations for the CPTs in the learning process, the identification of the benefits of using these representations and the empirical validation of our hypothesis. As mentioned in the introduction, we are not the first to consider efficient representations for the CPTs in the context of learning. Yet, to the best of our knowledge we are the first to consider and demonstrate the effects that these representations may have on the learning of the global structure of the network.

In addition, it is important to distinguish between the local representations we examine in this paper and the noisy-or and logistic regression models that have been examined in the literature. Both noisy-or and logistic regression (as applied in the Bayesian network literature) attempt to estimate the CPT with a *fixed* number of parameters. This number is usually linear in the number of parents in the CPT. In cases where the target distribution does not satisfy the assumptions embodied by these models, the estimates of CPTs produced by these methods can arbitrarily diverge from the target distribution. On the other hand, our local representations involve learning the structure of the CPT, which can range between a lean structure with few parameters to a full structure with an exponential number of parameters. Thus, our representations can scale up according to the complexity of the training data. This ensures that, in theory, they are asymptotically correct: given enough samples they will construct a close approximation of the target distribution.

We have focused our investigations on two fairly simple structured representations–trees and default tables. There are certainly many other possible representation of CPTs based for example on decision graphs, rules, CNF formulas, etc. (see [Boutilier, Friedman, Goldszmidt, and Koller 1996]). Our choice here was mainly due to the availability of efficient computational tools for learning the representations we use. The refinement of these methods deserves further attention. There are various approaches for learning trees in the Machine Learning literature, all of which can be easily incorporated in the learning procedures for Bayesian networks. In addition, there are possible interactions among the search procedures for global and local structures. These interactions can be exploited to reduce the computational cost of the learning process. We leave these issues for future research.

In conclusion, we have shown that the induction of local structured representation for CPTs significantly improves the performance of learning Bayesian networks. In essence, this is due to the fact that we have changed the *bias* of the scoring metric in the learning procedure to reflect the nature of the distribution in the data more accurately. Our experimental results show that networks learned using these local structured representations encode parameters that are more robust, and the induced distributions converge faster to the original distribution.

---

[5]We only show the results for a couple of such experiments. Similar qualitative behavior appears in all other experiments we generated.

### Acknowledgments

The authors are grateful to Wray Buntine and David Heckerman for comments on a previous draft of this paper and useful discussions relating to this work. Thanks also to Mark Peot of KI industries for providing the CTS and TJ networks. Parts of this work were done while Nir Friedman was at Rockwell Science Center. Nir Friedman was also supported in part by an IBM Graduate fellowship and NSF Grant IRI-95-03109.

## A  Bayesian Learning of Local Structured Representations

The MDL principle provided a straightforward framework for adjusting the metric to account for the additional structure in the representation of the CPTs. Another popular scoring metric for learning Bayesian networks is the Bayesian based metric described by Heckerman, Geiger and Chickering (HGC) [Heckerman, Geiger, and Chickering 1995] (which is based on earlier work of [Cooper and Herskovits 1992; Buntine 1991b]). We now proceed to sketch a similar modification of this metric deferring an in-depth treatment to the full version of the paper.

The Bayesian metric estimates the posterior probability of each network structure given the data. Learning amounts to searching for the network(s) that maximizes this probability. Let $G^h$ denote the hypothesis that the network structure is $G$, and let $\Theta_G$ represent the vector of parameters for structure $G$. The posterior probability we are interested in is $\Pr(G^h|D)$. Using Bayes rule we write this term as:

$$\Pr(G^h|D) = \alpha \Pr(D|G^h) \Pr(G^h)$$

where $\alpha$ is a normalization constant. The term $\Pr(G^h)$ is the prior probability on the network structure, and the term $\Pr(D|G^h)$ is the probability of the data given that the network structure is $G^h$. To evaluate the later term we must consider all possible parameter assignments to $G$. Thus:

$$\Pr(D|G^h) = \int \Pr(D|\Theta_G, G^h) \Pr(\Theta_G|G^h) d\Theta_G \quad (3)$$

where $\Pr(D|\Theta_G, G^h)$ is defined by Equation 1, and $\Pr(\Theta_G|G^h)$ is the prior density over parameter assignments to $G$. HGC (following [Cooper and Herskovits 1992]) identify a set of assumptions that justify decomposing this integral. Roughly speaking, they assume that each distribution $\theta_{X|\Pi_x}$ can be learned independently of all other distributions. Given this assumption they rewrite $\Pr(D|G^h)$ as:

$$\prod_i \prod_{\Pi_{x_i}} \int \prod_{x_i} \theta_{x_i|\Pi_{x_i}}^{N \cdot \hat{P}_D(x_i, \Pi_{x_i})} \Pr(\Theta_{X_i|\Pi_{x_i}}|G^h) d\Theta_{X_i|\Pi_{x_i}}$$

(4)

(This decomposition is analogous to the decomposition in Equation 2.) When the prior on each multinomial distribution $\Theta_{X_i|\Pi_{x_i}}$ is assumed to be a *Dirichlet prior*, the integrals in Equation 4 have a closed form solution [Heckerman 1995]). Roughly speaking, the prior density of the form *Dirichlet*$(\theta, k)$ is defined by two parameters, $\theta$ the expected value of the distribution of $X$, and $k$ the *equivalent sample size* which represents the confidence in the estimate.

There still remains a problem with the direct application of this method. For each possible network structure we would have to assign a prior on the parameter values. This is clearly infeasible since the number of possible structures is extremely large. HGC propose a set of assumptions that justify a method by which given a prior network $B^p$ and an equivalent sample size $N'$, we can assign prior probabilities to parameters in every possible network structure. Roughly speaking, the prior assigned to $\Theta_{X_i|\Pi_{x_i}}$ in a structure $G$ is computed from the prior distribution represented in $B^p$:

$$\Pr(\Theta_{X_i|\Pi_{x_i}}|G^h) \sim Dirichlet(\hat{\Theta}_{X_i|\Pi_{x_i}}, N'_{\Pi_{x_i}}),$$

where $\hat{\Theta}_{X_i|\Pi_{x_i}} = P_{B^p}(X_i|\Pi_{x_i})$ and $N'_{\Pi_{x_i}} = N' \cdot P_{B^p}(\Pi_{x_i})$. (Note that $\Pi_{X_i}$ are the parents of $X_i$ in $G$, but not necessarily in $B^p$.) Thus, their proposal essentially uses the conditional probability of $X_i$ given $\Pi_{x_i}$ in the prior network $B^p$ as the expected probability. Similarly, the equivalent sample size is taken to be proportional to the expected number of occurances of the values of $\Pi_{x_i}$.

We now sketch a proposal for a similar machinery that will enable the proper scoring of local structured representations. We denote by $L_G^h$ the hypothesis that $G$ has a local structure $L$ (which can be trees, default tables, or any other possible representations [Boutilier, Friedman, Goldszmidt, and Koller 1996]). We will also denote by $\Gamma_{X_i}^L$ the random variable associated with the local representation of the CPT of $X_i$.

We now write:

$$\Pr(G^h, L_G^h|D) = \alpha \Pr(D|L_G^h, G^h) \Pr(L_G^h|G^h) \Pr(G^h)$$

Specification of priors on local structures is a relatively simple problem, with no more complications than the specification of priors for the structure of the network $G^h$. Buntine [1991a, 1993], for example, suggests several possible priors on decision trees. A natural prior over local structures is defined using the MDL description length we described above, by setting $\Pr(L_G|G) = \alpha 2^{-DL(L_G)}$.

For the term $\Pr(D|L_G^h, G^h)$, we make an assumption similar to the one made by HGC (and by Buntine [1991b]): the parameter values for each possible value of the characteristic variable are independent. Thus, each multinomial sample is independent of the others, and we can derive the analogue of Equation 4 for $\Pr(D|L_G^h, G^h)$:

$$\prod_i \prod_{v \in Val(\Gamma_i^L)} \int \prod_{x_i} \theta_{x_i|v}^{N \cdot \hat{P}_D(x_i, v)} \Pr(\Theta_{X_i|v}|L_G^h, G^h) d\Theta_{X_i|v}$$

(5)

(This decomposition is analogous to the ones described in Propositions 3.1 and 3.2.) Again we assume that the priors $\Pr(\Theta_{X_i|v}|L_G^h, G^h)$ are Dirichlet, and thus the integrals have a closed form.

Once more we are faced with the problem of specifying a multitude of priors, that is, specifying $\Pr(\Theta_{X_i|v}|L_G^h, G^h)$ for each combination of possible local and global structures. Our objective, as in the case where the CPT is represented by a naive tabular form, is to set these priors from a prior distribution represented by a specific network $B^P$. We make two assumptions.

First, the prior for an instantiation of the characteristic variable does not depend on the structure of the representation. That is, a partition of the values of the parents of a node $X$ in the network only depends on the event that corresponds to this instantiation. For example, consider two possible trees for the same CPT, one that tests first on $Y$ and then on $Z$, and another that tests first on $Z$ and then on $Y$. Our assumption requires that the leaves that correspond to $Y = y, Z = z$, be assigned the same prior in both trees.

Second, we assume that the prior for a (larger) partition that corresponds to a union of several smaller partitions in another local structure is simply the weighted average of the smaller partitions. Once more consider two trees, one that consists of a single leaf, and another that has one test at the root. This assumption requires that the expected value of the parameters for the leaf in the first tree is the weighted avaerge of the expected values in the leaves of the second tree.

These assumptions follow directly from the assumption of equivalent sample size, which in the case of a naive unstructured representations follows from the HGC set of assumptions:[6] The assessement provided on the priors is equivalent to having started from complete ignorance, and seeing $N'$ cases of samples $D' = \{\mathbf{u}'_1, \ldots, \mathbf{u}'_{N'}\}$. Moreover, $P_{B^P}$, the probability represented in the prior network, describes the relative frequency of events among these samples, i.e., $P_{B^P} = \hat{P}_{D'}$. This assumption (combined with the appropriate assumptions from HGC) can be now used to derive the prior for $\Theta_{X_i|v}$ from a prior network as follows:

$$\Pr(\Theta_{X_i|v}|L_G^h, G^h) \sim Dirichlet(\hat{\Theta}_{X_i|v}, N'_v),$$

where $\hat{\Theta}_{X_i|v} = P_{B^P}(X_i|\Gamma_i^L = v)$ and $N'_v = N' \cdot P_{B^P}(\Gamma_i^L = v)$.

It remains to be seen how this Bayesian scoring metric for learning local structures performs in practice. We suspect that it would lead to improvements similar to those we observed for in MDL score. This intuition is based on the result by Schwarz [1978] which establishes that the two are essentially the same for sufficiently large $N$. Thus, if we set the priors for the structures such that $\log \Pr(G) = -DL_{graph}(G)$ (prior of the network structure is equal to its description length) and $\log \Pr(L_i^h) = -DL_L(L_i)$ where $DL_L$ is the appropriate description length function (e.g., $DL_T$), then Schwarz's result implies that that $\log \Pr(G, L|D) = -DL(G, L, D) + O(1)$. That is, for sufficiently large $N$ the two scores are essentially the same. Somewhat more complex arguments show that even for small samples, the scores are close when we start with the uninformative prior, i.e., one where $N' = 0$.

---

[6]We are grateful to David Heckerman for suggesting this simplifying assumption.